# High-Accuracy Prediction of Metal-Insulator-Metal Metasurface with Deep Learning

Kaizhu Liu, Hsiang-Chen Chui, Changsen Sun, and Xue Han[a]

**AFFILIATIONS**

School of Optoelectronic Engineering and Instrumentation Science, Dalian University of Technology, Dalian 116024, China

[a] Author to whom correspondence should be addressed: xue_han@dlut.edu.cn

Abstract:

Deep learning prediction of electromagnetic software calculation results has been a widely discussed issue in recent years. But the prediction accuracy was still one of the challenges to be solved. In this work, we proposed that the ResNets-10 model was used for predicting plasmonic metasurface $S_{11}$ parameters. The two-stage training was performed by the k-fold cross-validation and small learning rate. After the training was completed, the prediction loss for aluminum, gold, and silver metal-insulator-metal metasurfaces was -48.45, -46.47, and -35.54, respectively. Due to the ultralow error value, the proposed network can replace the traditional electromagnetic computing method for calculation within a certain structural range. Besides, this network can finish the training process less than 1,100 epochs. This means that the network training process can effectively lower the design process time. The ResNets-10 model we proposed can also be used to design meta-diffractive devices and biosensors, thereby reducing the time required for the calculation process. The ultralow error of the network indicates that this work contributes to the development of future artificial intelligence electromagnetic computing software.

In recent years, many new phenomena have been revealed by using micro-nano structures to regulate electromagnetic waves. The researchers developed micro-nano devices [1] by utilizing periodic arrangements in two-dimensional (2D) zones. Especially metasurfaces, 2D artificial nanostructures [2], were designed and fabricated to control the propagation of electromagnetic waves and resulted in the unique optical properties beyond traditional materials. Metallic materials were used to design and manufacture chiral [3] and polarization-independent [4] metasurfaces within the visible light range. With the surface plasmon resonance (SPR) effect, the proposed metasurfaces can be contrived for surface enhanced Raman spectroscopy (SERS) [5], optical filters [6], chemical sensors [7], and biosensors [8]. To adjust the phase and intensity of every unit on the planned metasurface can realize lots of optical devices, such as metalens [9, 10], holograms [11], and metagratings [12]. Nano-photonics devices have complex structures and unique optical properties, requiring comprehensive consideration of multiple factors to design excellent devices. Traditional calculation methods were seen inefficient in the design of nano-photonics devices, due to requiring a lot of times and efforts [13]. Using deep learning for nano-photonics device design, one can speed up the design process, improve design accuracy, and reduce design costs.

In the field of metasurface computing, there have been a number of advancements in predicting and inverse designing the performance of metasurfaces [14, 15]. In 2019, Sensong et al. [16] used neural tensor network (NTN) to achieve forward and backward prediction of amplitude and phase for dielectric metasurfaces in the 5-10 μm wavelength range. Using convolutional neural networks to process unit structure images, so that the structural is not limited to a fixed form and can be any forms. In the same year, Christian et al. [17] predicted the transmittance of metasurfaces consisting of arbitrarily combined media with different heights and radii with using a fully connected layers' neural network structure. In 2022, Zhang et al. [18] used transfer-learning to reduce the amount of training data. They achieved good results by trying three different structural scenarios, each with different parameterizations, different physical sizes, and completely different geometries. This work has made significant contributions to reducing the accumulation of preliminary data. In addition, Omar et al. [19] made efforts on the interpretability of neural networks in the deep learning prediction of metasurfaces. They introduced the Drude-Lorentz model to describe the optical constants and integrated it into the neural network calculation, which makes the network no longer a black-box computation. Recently, Sensong et al. [20] combined NTN and convolutional layers to accomplish the prediction of Scattering (S) parameters of $Ge_2Sb_2Se_4Te$ under different crystallization degrees and amorphous states. The above work has made contributions in improving the structural degrees of freedom and neural network interpretability in deep learning for predicting the performance of metasurface. Table 1 contains the neural network test loss of these works, which is an important indicator for the practicality of the neural network. The forward predictive test loss was roughly between $3\times10^{-3}$ and $2\times10^{-4}$, indicating that there still exists a considerable amount of loss. Developing an artificial intelligence-based computing method to replace traditional and complex electromagnetic software requires addressing the crucial issue of reducing computational error. Therefore, further research on selecting appropriate neural network structures or adopting methods that combine physical information to reduce loss in similar tasks remains a crucial issue that needs to be studied.

**Table 1.** The forward prediction metasurface performance test loss

| Works | Test loss | Activation function | Loss function |
|---|---|---|---|
| Ref. 16 | $2.9\times10^{-4}$ | ReLu | MSE |
| Ref. 17 | $1.1\times10^{-3}$ | ReLu | MSE |
| Ref. 18 | $3.0\times10^{-3}$ | ReLu | MSE |
| Ref. 19 | $1.5\times10^{-3}$ | - | MSE |
| Ref. 20 | $7.5\times10^{-4}$ | ReLu | MSE |
| This work | Al-ResNets-10: $1.4\times10^{-5}$<br>Au-ResNets-10: $2.6\times10^{-5}$<br>Ag-ResNets-10: $2.8\times10^{-4}$ | Mish | SmoothL1 |

In our previous work [21], we used a specially designed complex residual neural network (CRNN) to accurately predict the $S_{21}$ Parameters of dielectric metasurfaces in the near-infrared range. We found that the CRNN could predict data with gentle or few peak changes quite well, but could not respond to data with multiple dense peaks effectively. SPR metasurfaces were commonly designed with precious metal materials such as gold (Au) and silver (Ag) in the visible-near-infrared range. Aluminum (Al), as an inexpensive metal, has also gradually received attention in recent years due to its low price [22, 23]. Recent research has shown that the resonance modes presented by metasurfaces based on the SPR effect are typically broadband resonances, compared to dielectric metasurfaces [24,25]. This indicates that the CRNN network architecture we proposed may be suitable for predicting the $S_{11}$ parameters of metasurfaces based on the SPR effect. Therefore, we proposed that the ResNets-10 model to predict the metal-insulator-metal (MIM) metasurface. The MIM metasurface forward prediction work was carried out using the three common metals, i.e., Au, Ag, and Al.

We used the finite difference time domain (FDTD) method to calculate the $S_{11}$ parameter of MIM metasurfaces constructed from Au, Ag, and Al, within the wavelength range between 500 to 850 nm. The background refractive index was set 1.33. The isolation layer was silicon dioxide ($SiO_2$), and the reflection layer was Al. Fig. 1(a) showed the MIM metasurface structure. Our datasets were obtained by setting Au, Ag, and Al nanostructures based on four MIM metasurface parameters (height (H), period (P), radius (R), and thickness (T)) with proper gradient arrangement. Each metal datasets size consisted of 6561 pieces inside one data. The detailed information of the related dataset was presented in the supplementary information.

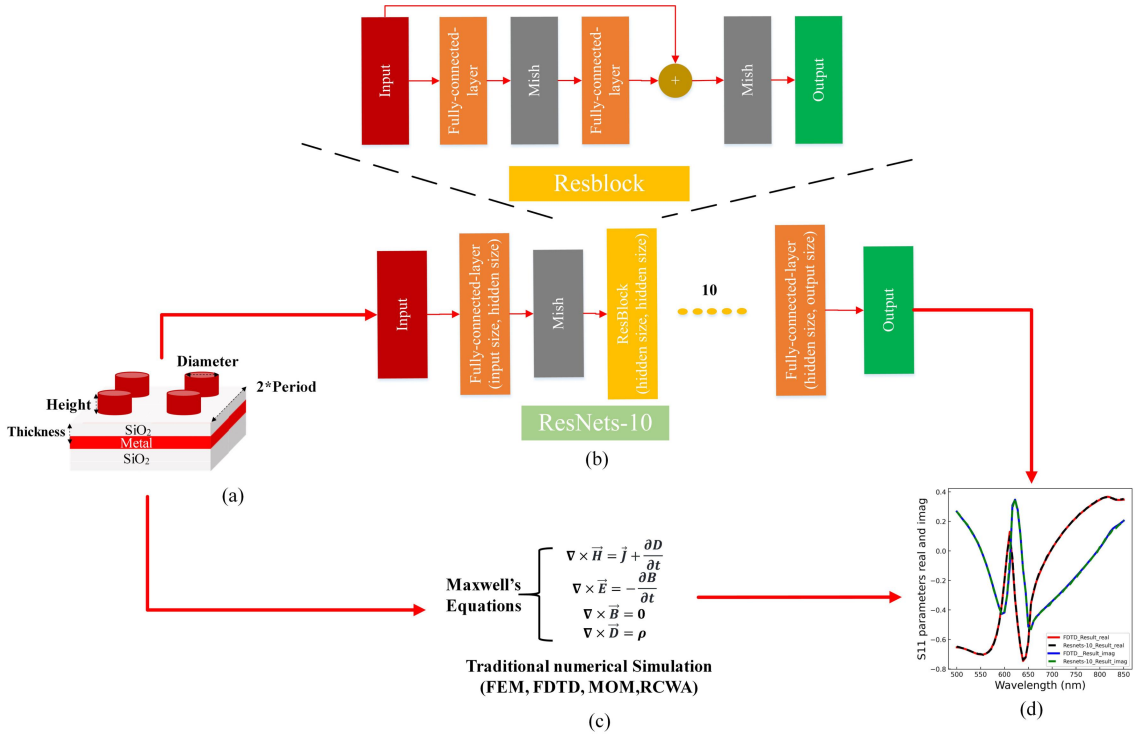

Fig. 1 Description of modules of the ResNets-10 model compose and function. MIM metasurface structure diagram. (b) ResNets-10 model composed. (c) traditional numerical simulation methods. (d) $S_{11}$ parameters by ResNets-10 model prediction and FDTD calculation.

The ResNets-10 model was built under the Windows 11 operating system. The laptop was configured with Intel Core i7-12700H CPU at 2.6 GHz to 4.7 GHz/8 GB RTX 3070/16 GB Memory/1 T SSD. The deep learning algorithm was implemented on the Anaconda platform. The Python version is 3.8.8, and the Pytorch version is 1.8.0. In comparison to the conventional ResNets model approach, we used fully connected layers in place of convolutional layers for internal connections. The structural and propagation modes of ResNets-10 model are shown in Fig. 1(b). The traditional calculation methods, such like finite element method (FEM), FDTD, and so on, were shown in Fig. 1(c). Comparing with traditional calculation methods, the neural network learns rules from datasets to predict results instead of relying on theoretical deductions. It can substantially boost the computational speed. As depicted in Fig. 1 (d), the ultimate objective was for the prediction results from neural network to align with the calculation results of the traditional software with a high degree of fidelity. The activation function was Mish. The size of the hidden layer in the residual block (Resblock) was 256, while the output size was 64. The input size was 4, there had four parameters H, P, R, and T, respectively.

For MIM metasurface units, we used parallel training with ResNets-10 model to predict both the real and imaginary part of the $S_{11}$. The SmoothL1 function was chosen for loss function. The optimizer was Adam. The dataset normalization method was referenced from our previous work [18]. The training process of ResNets-10 was presented in Fig. 2. Our datasets to each metal only contain 6561 pieces of data, more data we acquired can be used to training and make sure the train result performances can be better. We used k-fold cross-validation to train the model and randomly selected 609 pieces of data for the test dataset. Unlike traditional k-fold cross-validation, each folded model is inherited and not recalculated. This purpose was to prevent the model from overfitting. When a low validation error was achieved during k-fold cross-validation, it indicates that it was possible to proceed to the next training step. The usability of the ResNets-10 model was evaluated by observing the training and testing loss. The datasets of Au, Ag, and Al within the 500 to 850 nm wavelength range were similar. We can train the $S_{11}$ parameters for one material to predict the nanostructures initially. Then, we can design the nanostructures of other materials with transfer learning. It can accelerate the convergence speed and enhance prediction performance. We evaluated the data, and the training and testing loss for Al MIM metasurface were the best. Therefore, we started training using Al MIM metasurface.

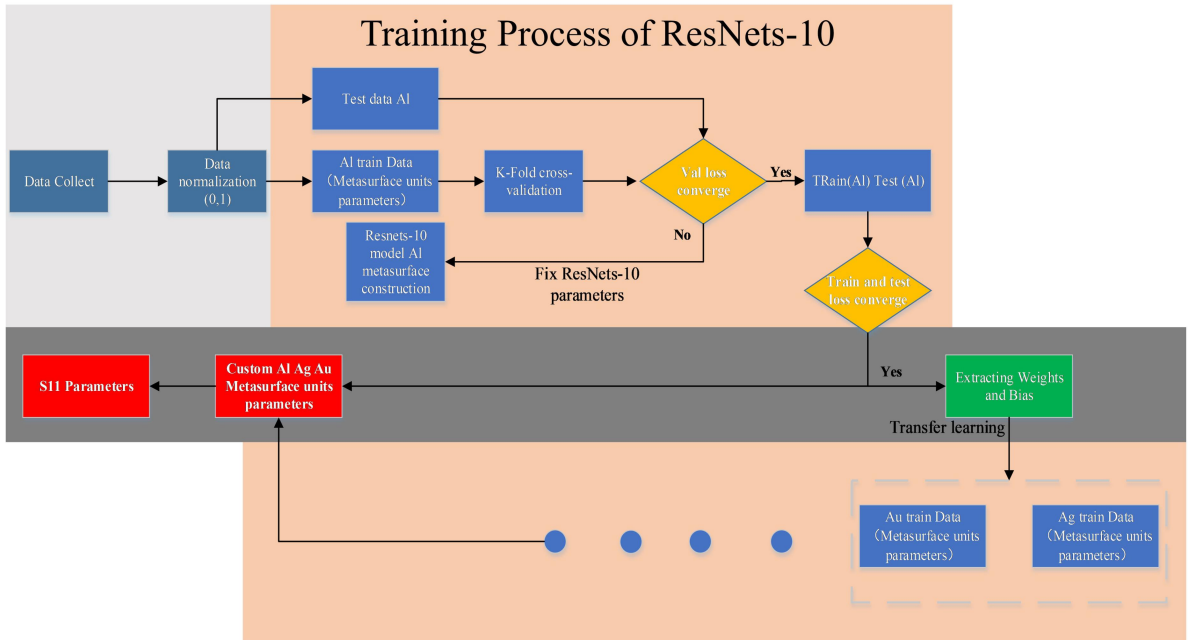

Fig. 2 Training process of ResNets-10 model's training and prediction process.

In the k-fold cross-validation, the k factor was set as 10 to train the Al MIM metasurface. The training process for each fold consisted of 100 epochs, with an initial learning rate of $5\times10^{-4}$ and a batch size of 128. After the 10-fold cross-validation training process, the loss of Al-Resnets-10 model in validation was calculated as -46.5. It means the model pretraining was success. Therefore, we used the Al-ResNets-10 model weights and bias transfer to run k-fold cross-validation for both the Au-ResNets-10 model and Ag-ResNets-10 model. Since the Ag $S_{11}$ parameters compared Au and Al had more resonance peak, we use small learning rate of $3\times10^{-4}$ to ensure the k-fold cross-validation process can be better progress. It can be seen in Fig. 3 (a) that the 1-fold cross-validation loss for Ag and Au was well below that of Al, which means that the transfer learning in this work has significant assistance. In Fig. 3 (b), we present the train and test loss of Au-ResNets-10 model after 5-fold cross-validation. Figures S1(a) and S1(b) showed the Al and Ag train and test losses in the supplementary information. The train and test losses were enough low compared to current literature reports. The validation loss, train loss, and test loss of Al, Au and Ag MIM metasurfaces were indicated in Table 2.

**Table 2.** The validation loss, train loss, and test loss of Al, Au and Ag MIM metasurfaces

| Materials | Validation loss | Train loss | Test loss |
|---|---|---|---|
| Al | -46.46 | -53.10 | -48.45 |
| Au | -47.46 | -53.59 | -46.47 |
| Ag | -42.65 | -50.05 | -35.54 |

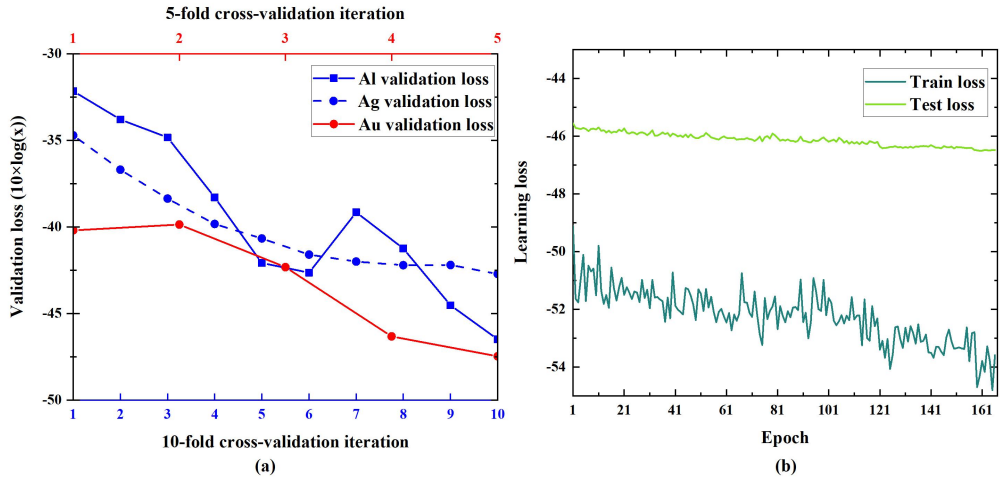

Fig. 3 Training Resnets-10 model process. (a) k-fold cross-validation iteration loss. The learning rates for Al and Au were both $5\times10^{-4}$, while that of Ag was $3\times10^{-4}$. (b) Au MIM metasurface units train and test loss. The learning rate was $1\times10^{-4}$.

We present the results of the test dataset in Fig. 4 (more test results see supplementary information Figure S2, Figure S3 and Figure S4), where the solid red and blue lines correspond to the $S_{11}$ parameters that were calculated using FDTD, and the dashed black and green lines represent the predicted $S_{11}$ parameters by the ResNets-10 model. The test results show that ResNets-10 model can achieve accurate predictions of the $S_{11}$ parameter for MIM metasurface structures composed of three different materials, Al, Au, and Ag. Among them, Al MIM metasurface units perform best in this network structure, and the test loss reached -48.45. Even Ag MIM metasurface units, the worst performing material, reached a loss of -35.54, which was comparable to the reported losses that have been achieved by deep learning studies so far. The prediction loss was found to be low, indicating that the ResNets-10 model can replace the calculation work for Al, Au, and Ag at this wavelength range. This substitution can significantly reduce the calculation time required to design an MIM metasurface at a wavelength range of 500 to 850 nm.

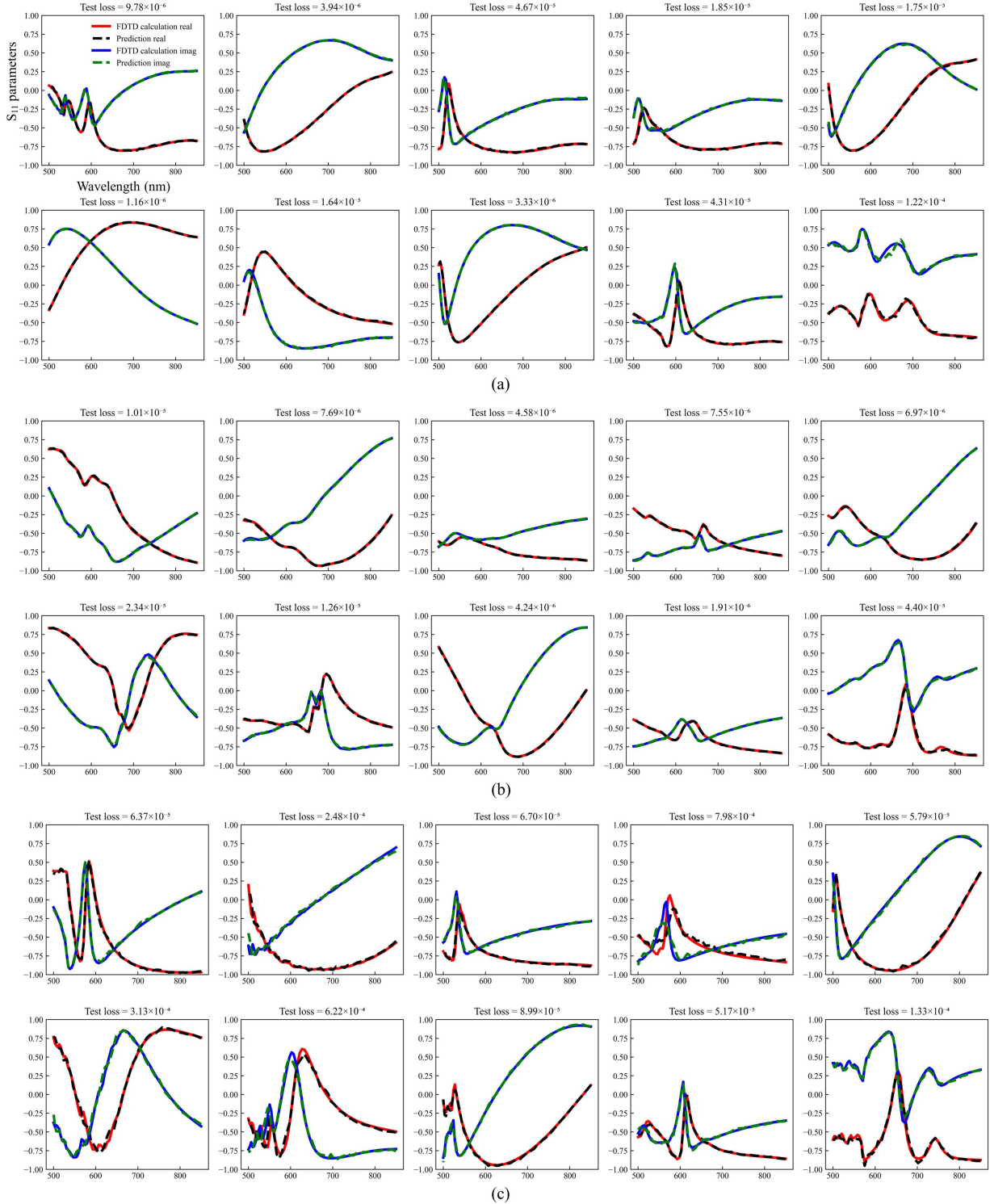

Fig. 4 The ResNets-10 model for test datasets prediction results (a) Al MIM metasurface units (b) Au MIM metasurface units (c) Ag MIM metasurface units. The red and blue solid lines represent

the real and imaginary parts respectively of the FDTD calculated $S_{11}$ parameter. The red and blue dashed lines represent the real and imaginary parts respectively of the ResNets-10 model predicted $S_{11}$ parameter.

For example, the calculation of one group Au MIM metasurface units phase change at 788 nm wavelengths was showed in Fig. 5(a). The FDTD calculation data was served as ground truth, and the ResNets-10 model prediction data was compared. The H was 30 nm, S was 300 nm and T was 80 nm. The R was set from 30 to 100 nm with the gradient differences of 5 nm. Two data curves from the FDTD calculation and the ResNets-10 model prediction were near perfectly matching. By utilizing the phase change of this unit, the next step of designing meta-diffractive devices such as metalens and metagrating can be achieved. We can also make predictions for intensity-related work. Take it as another example, SERS with 532 nm laser by SPR effect usually need the structured resonance closed to the incidence laser wavelength. Optimization of structural parameters was needed to realize this purpose. Scanning multiple sets of Al MIM metasurface units can obtain the $S_{11}$ parameters, as shown in Fig. 5(b). The H was 20 nm, S was 370 nm and T was 82.5 nm. The R was set from100 to 130 nm with the gradient differences of 5 nm. It can be known from the prediction results that when R was 125 μm, the resonance peak wavelength position was 532 nm.

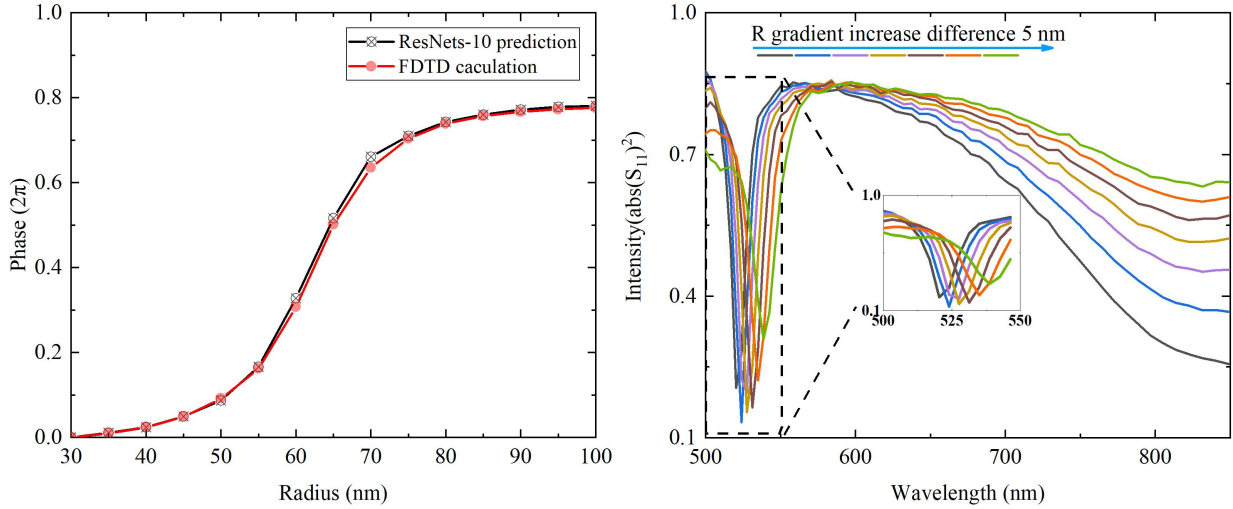

Fig. 5 The ResNets-10 model for simulation work results (a) Au MIM metasurface units phase change with R at 30 to 100 nm changes (b) Al MIM metasurface units $S_{11}$ intensity with R at 100 to 150 nm changes.

In conclusions, we realized the ultralow losses by using the ResNets-10 model. This approach can further improve the accuracy of neural network forward prediction of MIM metasurface $S_{11}$ parameters. For Al MIM metasurfaces, a test error of approximately -48.45 was measured. The training can be completed after approximately 100 epochs following k-fold cross-validation. And the use of transfer learning techniques has improved the predictive performance for different materials. Compared to the recently reported work, only 10% of the number of epochs is required for training. The ResNets-10 model can also assist in exploring and designing structural parameters for meta-diffractive devices and SERS. The time required to complete regular parameter scans was almost instantaneous, it greatly reduces the time and costs associated with trial and error for MIM metasurface. The ReNets-10 can also be extended to SPR effect structures that do not have reflective layers, as well as to designing with more metal materials. Due to its ultralow loss characteristics, this proposed approach is expected to contribute to the development of artificial intelligence electromagnetic calculation software in the future.

**Supplementary Material**

For information on the dataset, training hyperparameters of the ResNets-10 model, training process, and more testing data, please refer to the supplementary materials.

**Acknowledgment**

This work was supported by the National Natural Science Foundation of China (Grant Nos. 62105053).

**AUTHOR DECLARATIONS**

**Conflict of Interest**

The authors have no conflicts to disclose.

**Author Contributions**

**Kaizhu Liu:** Conceptualization (lead); Investigation (lead); Methodology (lead); Visualization (lead); Writing – original draft (lead).

**Hsiang-Chen Chui:** Conceptualization (equal); Visualization (equal); Writing – review & editing (equal).

**Changsen Sun:** Funding acquisition (equal); Resources (equal); Supervision (equal).

**Xue Han:** Funding acquisition (equal); Resources (equal); Supervision (equal).

## DATA AVAILABILITY

The data that support the findings of this study are available within the article and its supplementary information.

## REFERENCE

1. Pengfei Qiao, Weijian Yang, and Connie J. Chang-Hasnain, Adv. Opt. Photonics 10 (1) (2018).
2. Y. Nanfang, P. Genevet, F. Aieta, M. A. Kats, R. Blanchard, G. Aoust, J. P. Tetienne, Z. Gaburro, and F. Capasso, IEEE J. Sel. Top. Quantum Electron. 19 (3), 4700423 (2013).
3. Y. Zhao, A. N. Askarpour, L. Sun, J. Shi, X. Li, and A. Alu, Nat. Commun. 8, 14180 (2017).
4. F. Jiao, F. Li, J. Shen, C. Guan, S. A. Khan, J. Wang, Z. Yang, and J. Zhu, Sens. Actuators, B 344 (2021).
5. G. Barbillon, Int. J. Mol. Sci. 23 (18) (2022).
6. Y. S. Lin, J. Dai, Z. Zeng, and B. R. Yang, Nanoscale Res. Lett. 15 (1), 77 (2020).
7. S. Lin, W. Liu, X. Hou, Z. Peng, Z. Chen, and F. Hu, Spectrochim. Acta, Part A 292, 122413 (2023).
8. Peiliang Wang, Jing Lou, Yun Yu, Lang Sun, Lan Sun, Guangyou Fang, and Chao Chang, Nano Res. (2023).
9. S. Wang, P. C. Wu, V. C. Su, Y. C. Lai, C. Hung Chu, J. W. Chen, S. H. Lu, J. Chen, B. Xu, C. H. Kuan, T. Li, S. Zhu, and D. P. Tsai, Nat. Commun. 8 (1), 187 (2017).
10. Sajan Shrestha, Adam Overvig, Ming Lu, Aaron Stein, and Nanfang Yu, Appl. Phys. Lett. 122 (20) (2023).
11. Zi. Deng, M. Jin, X. Ye, S. Wang, T. Shi, J. Deng, N. Mao, Y. Cao, B. Guan, A., G. Li, and X. Li, Adv. Funct. Mater. 30 (21) (2020).
12. S. Gao, W. Yue, C. Park, S. Lee, E. Kim, and D. Choi, ACS Photonics 4 (2), 322 (2017).
13. W. Ma, Z. Liu, Z. A. Kudyshev, A. Boltasseva, W. Cai, and Y. Liu, Nat. Photonics 15 (2), 77 (2020).
14. Nathan Bryn Roberts and Mehdi Keshavarz Hedayati, Appl. Phys. Lett. 119 (6) (2021).
15. Xiao Qing Chen, Lei Zhang, and Tie Jun Cui, Appl. Phys. Lett. 122 (16) (2023).
16. S. An, C. Fowler, B. Zheng, M. Y. Shalaginov, H. Tang, H. Li, L. Zhou, J. Ding, A. M. Agarwal, C. Rivero-Baleine, K. A. Richardson, T. Gu, J. Hu, and H. Zhang, ACS Photonics 6 (12), 3196 (2019).
17. C. C. Nadell, B. Huang, J. M. Malof, and W. J. Padilla, Opt. Express 27 (20), 27523 (2019).
18. J. Zhang, C. Qian, Z. Fan, J. Chen, E. Li, J. Jin, and H. Chen, Adv. Opt. Mater. 10 (17) (2022).
19. O. Khatib, S. Ren, J. Malof, and W. J. Padilla, Adv. Opt. Mater. 10 (13) (2022).
20. S. An, B. Zheng, M. Julian, C. Williams, H. Tang, T. Gu, H. Zhang, H. Kim, and J. Hu, Nanophotonics 11 (17), 4149 (2022).

21  K. Liu and C. Sun, Appl. Opt. 62 (5), 1200 (2023).
22  R. H. Siddique, S. Kumar, V. Narasimhan, H. Kwon, and H. Choo, ACS Nano 13 (12), 13775 (2019).
23  Y. Zhang, L. Shi, D. Hu, S. Chen, S. Xie, Y. Lu, Y. Cao, Z. Zhu, L. Jin, B. Guan, S. Rogge, and X. Li, Nanoscale Horiz. 4 (3), 601 (2019).
24  L Aleksandrs, T Andreas, L Ming, and L Bang, Sci. Adv. 5 (5) (2019).
25  A. V. Kabashin, P. Evans, S. Pastkovsky, W. Hendren, G. A. Wurtz, R. Atkinson, R. Pollard, V. A. Podolskiy, and A. V. Zayats, Nat. Mater. 8 (11), 867 (2009).